\documentclass[runningheads]{llncs}
\usepackage{graphicx}
\usepackage{amsmath,amssymb} 
\usepackage{color}
\usepackage{xspace}

\usepackage{pgfplots}
\usepackage{pgfplotstable}
\usepackage{pgf,tikz}
\usetikzlibrary{arrows,shapes,decorations.markings}
\pgfplotsset{compat=1.9}

 \usepgfplotslibrary{patchplots}

\newcommand{\extdata}[1]{\input{#1}}

\newcommand{\leg}[1]{\addlegendentry{#1}}

\newcommand{\head}[1]{{\smallskip\noindent\bf #1}}

\newcommand{\real}{\mathbb{R}}
\def\l2{\ensuremath{\ell_2}\xspace}

\newcommand{\T}{{\!\top}}

\newcommand{\diag}{\operatorname{diag}}
\newcommand{\defn}{\mathrel{\operatorname{:=}}}
\newcommand{\norm}[1]{\left\|{#1}\right\|}

\newcommand{\gen}[1]{\left\langle{#1}\right\rangle}

\def\ssp{\hspace{3pt}}
\def\msp{\hspace{5pt}}

\newcommand{\cK}{\mathcal{K}}
\newcommand{\cL}{\mathcal{L}}

\newcommand{\cW}{\mathcal{W}}

\newcommand{\vO}{\mathbf{O}}

\newcommand{\vb}{\mathbf{b}}

\newcommand{\ve}{\mathbf{e}}

\newcommand{\vv}{\mathbf{v}}

\newcommand{\vx}{\mathbf{x}}
\newcommand{\vy}{\mathbf{y}}
\newcommand{\vz}{\mathbf{z}}

\newcommand{\vone}{\mathbf{1}}

\def\roxf{$\mathcal{R}$Oxford\xspace}
\def\rox{$\mathcal{R}$Oxf\xspace}

\def\rpar{$\mathcal{R}$Paris\xspace}
\def\rpa{$\mathcal{R}$Par\xspace}

\def\r1m{$\mathcal{R}$1M\xspace}

\def\ssp{\hspace{3pt}}
\def\msp{\hspace{5pt}}

\makeatletter
\DeclareRobustCommand\onedot{\futurelet\@let@token\@onedot}
\def\@onedot{\ifx\@let@token.\else.\null\fi\xspace}
\def\eg{\emph{e.g}\onedot} 
\def\ie{\emph{i.e}\onedot} 
 
\def\etc{\emph{etc}\onedot} \def\vs{\emph{vs}\onedot}
 
\def\etal{\emph{et al}\onedot}
\makeatother

\begin{document}

\title{Hybrid Diffusion: Spectral-Temporal Graph Filtering for Manifold Ranking} 

\titlerunning{Hybrid Diffusion}

\author{
Ahmet Iscen$^1$\ \ \ \ Yannis Avrithis$^2$\ \ \ \ Giorgos Tolias$^1$ \\ Teddy Furon$^2$ \ \ \ \ Ond{\v r}ej Chum$^1$
}

\authorrunning{A.Iscen, Y. Avrithis, G. Tolias, T. Furon and O. Chum}

\institute{$^1$VRG, FEE, CTU in Prague\ \ \ \ \ \ $^2$Univ Rennes, Inria, CNRS, IRISA}

\maketitle

\begin{abstract}
State of the art image retrieval performance is achieved with CNN features and manifold ranking using a $k$-NN similarity graph that is pre-computed off-line. The two most successful existing approaches are \emph{temporal filtering}, where manifold ranking amounts to solving a sparse linear system online, and \emph{spectral filtering}, where eigen-decomposition of the adjacency matrix is performed off-line and then manifold ranking amounts to dot-product search online. The former suffers from expensive queries and the latter from significant space overhead. Here we introduce a novel, theoretically well-founded \emph{hybrid filtering} approach allowing full control of the space-time trade-off between these two extremes. Experimentally, we verify that our hybrid method delivers results on par with the state of the art, with lower memory demands compared to spectral filtering approaches and faster compared to temporal filtering.
\end{abstract}


\section{Introduction}

Most image retrieval methods obtain their initial ranking of the database images by computing similarity between the query descriptor and descriptors of the database images. Descriptors based on local features~\cite{SZ03,PCISZ07} have been largely replaced by more efficient CNN-based image descriptors~\cite{GARL17,RTC18}.
Regardless of the initial ranking, the retrieval performance is commonly boosted by considering the manifold structure of the database descriptors, rather than just independent distances of query to database images.
Examples are query expansion~\cite{CPSIZ07,AZ12} and diffusion~\cite{DB13,ITA+17,IAT+18}. \emph{Query expansion} uses the results of initial ranking to issue a novel, enriched, query~\cite{CPSIZ07} on-line only. \emph{Diffusion} on the other hand, is based on the $k$-NN graph of the dataset that is constructed off-line, so that, assuming novel queries are part of the dataset, their results are essentially pre-computed. Diffusion can then be seen as infinite-order query expansion~\cite{ITA+17}.

The significance of the performance boost achieved by diffusion has been recently demonstrated at the ``Large-Scale Landmark Recognition''\footnote{\url{https://landmarkscvprw18.github.io/}} challenge in conjunction with CVPR 2018. The vast majority of top-ranked teams have used query expansion or diffusion
as the last step of their method.

Recently, efficient diffusion methods have been introduced to the image retrieval community. Iscen \etal~\cite{ITA+17} apply diffusion to obtain the final ranking, in particular by solving a large and sparse system of linear equations. Even though an efficient \emph{conjugate gradient} (CG)~\cite{Hack94} solver is used, query times on large-scale datasets are in a range of several seconds. A significant speed-up is achieved by \emph{truncating} the system of linear equations. Such an approximation, however, brings a slight degradation in the retrieval performance. Their method can be interpreted as graph
filtering in the {\em temporal} domain.

In the recent work of Iscen \etal~\cite{IAT+18}, more computation is shifted to the off-line phase to accelerate the query. The solution of the linear system is estimated by low-rank approximation of the $k$-NN graph Laplacian. Since the eigenvectors of the Laplacian represent a Fourier basis of the graph, this is interpreted as graph filtering in the {\em spectral} domain.
The price to pay
is increased space complexity to store the embeddings of the database descriptors. For comparable performance, a 5k-10k dimensional vector is needed per image.

In this paper, we introduce a \emph{hybrid} method that combines spectral filtering~\cite{IAT+18} and temporal filtering%
~\cite{ITA+17}. This hybrid method offers a trade-off between speed (\ie, the number of iterations of CG) and the additional memory required
(\ie, the dimensionality of the embedding). The two approaches~\cite{ITA+17,IAT+18} are extreme cases of our hybrid method. We show that the proposed method pairs or outperforms the previous methods while either requiring less memory or being significantly faster -- only three to five iterations of CG are necessary for embeddings of 100 to 500 dimensions.

While both temporal and spectral filtering approaches were known in other scientific fields before being successfully applied to image retrieval, to our knowledge the proposed method is novel and can be applied to other domains.

The rest of the paper is organized as follows. Related work is reviewed in Section~\ref{sec:related}. Previous work on temporal and spectral filtering is detailed in Sections~\ref{sec:temporal} and~\ref{sec:spectral} respectively, since the paper builds on this work. The proposed method is described in Section~\ref{sec:hybrid} and its behavior is analyzed in Section~\ref{sec:analysis}. Experimental results are provided in Section~\ref{sec:exp}. Conclusions are drawn in Section~\ref{sec:conclusions}.
\section{Related work}
\label{sec:related}
Query expansion (QE) has been a standard way to improve recall of image retrieval since the work of Chum \etal~\cite{CPSIZ07}.
A variety of approaches exploit local feature matching and perform various types of verification.
Such matching ranges from selective kernel matching~\cite{TJ14} to geometric consensus~\cite{CPSIZ07,CMPM11,JB09} with RANSAC-like techniques.
The verified images are then used to refine the global or local image representation of a novel query.

Another family of QE methods are more generic and simply assume a global image descriptor~\cite{SLBW14,JHS07,DJAH14,DGBQG11,ZYCYM12,DB13,AZ12}.
A simple and popular one is average-QE~\cite{CPSIZ07}, recently extended to $\alpha$-QE~\cite{RTC18}.
At some small extra cost, recall is significantly boosted.
This additional cost is restricted to the on-line query phase. This is in contrast to another family of approaches that considers an off-line pre-processing of the database.
Given the nearest neighbors list for database images, QE is performed by adapting the local similarity measure~\cite{JHS07}, using reciprocity constraints~\cite{DJAH14,DGBQG11} or graph-based similarity propagation~\cite{ZYCYM12,DB13,ITA+17}. The graph-based approaches, also known as diffusion, are shown to achieve great performance~\cite{ITA+17} and to be a good way for feature fusion~\cite{ZYCYM12}.
Such on-line re-ranking is typically orders of magnitude more costly than simple average-QE.

The advent of CNN-based features, especially global image descriptors, made QE even more attractive.
Average-QE or $\alpha$-QE are easily applicable and very effective with a variety of CNN-based descriptors~\cite{RTC18,GARL17,TSJ16,KMO15}.
State-of-the-art performance is achieved with diffusion on global or regional  descriptors~\cite{ITA+17}.
The latter is possible due to the small number of regions that are adequate to represent small objects, in contrast to thousands in the case of local features.

Diffusion based on tensor products can be attractive in terms of performance~\cite{BBT+18,BZW+17}. However, in this work, we focus on the page-rank like diffusion~\cite{ITA+17,ZWG+03} due to its reasonable query times. An iterative on-line solution was commonly preferred~\cite{DB13} until the work of Iscen \etal~\cite{ITA+17}, who solve a linear system to speed up the process.
Additional off-line pre-processing and the construction and storage of additional embeddings reduce diffusion to inner product search in the spectral ranking of Iscen \etal~\cite{IAT+18}. This work lies exactly in between these two worlds and offers a trade-off exchanging memory for speed and vice versa.

Fast random walk with restart~\cite{ToFP06} is very relevant in the sense that it follows the same diffusion model as~\cite{ITA+17,ZWG+03} and is a hybrid method like ours. It first disconnects the graph into distinct components through clustering and then obtains a low-rank spectral approximation of the residual error. Apart from the additional complexity, parameters \etc of the off-line clustering process and the storage of both eigenvectors and a large inverse matrix, its online phase is also complex, involving the Woodbury matrix identity and several dense matrix-vector multiplications. Compared to that, we first obtain a \emph{very} low-rank spectral approximation of the original graph, and then solve a sparse linear system of the residual error. Thanks to orthogonality properties, the online phase is nearly as simple as the original one and  significantly faster.
\section{Problem formulation and background}
\label{sec:background}

The methods we consider are based on a nearest neighbor graph of a dataset of $n$ items, represented by $n \times n$ \emph{adjacency matrix} $W$. The graph is undirected and weighted according to similarity: $W$ is sparse, symmetric, nonnegative and zero-diagonal. We symmetrically normalize $W$ as $\cW \defn D^{-1/2} W D^{-1/2}$, where $D \defn \diag(W \vone)$ is the diagonal \emph{degree matrix}, containing the row-wise sum of $W$ on its diagonal.
The eigenvalues of $\cW$ lie in $[-1,1]$~\cite{Chun97}.

At query time, we are given a sparse $n \times 1$ \emph{observation vector} $\vy$, which is constructed by searching for the nearest neighbors of a query item in the dataset and setting its nonzero entries to the corresponding similarities. The problem is to obtain an $n \times 1$ \emph{ranking vector} $\vx$ such that retrieved items of the dataset are ranked by decreasing order of the elements of $\vx$. Vector $\vx$ should be close to $\vy$ but at the same time similar items are encouraged to have similar ranks in $\vx$, essentially by exploring the graph to retrieve more items.

\subsection{Temporal filtering} \label{sec:temporal}

Given a parameter $\alpha \in [0,1)$, define the $n \times n$ \emph{regularized Laplacian} function by
\begin{equation}
	\cL_\alpha(A) \defn (I_n - \alpha A) / (1 - \alpha)
\label{eq:laplacian}
\end{equation}
for $n \times n$ real symmetric matrix $A$, where $I_n$ is the $n \times n$ identity matrix. Iscen \etal~\cite{ITA+17} then define $\vx$ as the unique solution of the linear system
\begin{equation}
	\cL_\alpha(\cW) \vx = \vy,
\label{eq:temporal}
\end{equation}
which they obtain approximately in practice by a few iterations of the \emph{conjugate gradient} (CG) method, since $\cL_\alpha(\cW)$ is positive-definite. At large scale, they truncate $\cW$ by keeping only the rows and columns corresponding to a fixed number of neighbors of the query, and re-normalize. Then,~(\ref{eq:temporal}) only performs re-ranking within this neighbor set.

We call this method \emph{temporal\footnote{``Temporal'' stems from conventional signal processing where signals are functions of ``time''; while ``spectral'' is standardized also in graph signal processing.} filtering} because if $\vx$, $\vy$ are seen as signals, then $\vx$ is the result of applying a linear graph filter on $\vy$, and CG iteratively applies a set of recurrence relations that determine the filter. While $\cW$ is computed and stored off-line,~(\ref{eq:temporal}) is solved online (at query time), and this is expensive.

\subsection{Spectral filtering} \label{sec:spectral}

Linear system~(\ref{eq:temporal}) can be written as $\vx = h_\alpha(\cW) \vy$, where the \emph{transfer function} $h_\alpha$ is defined by
\begin{equation}
	h_\alpha(A) \defn \cL_\alpha(A)^{-1} = (1 - \alpha) (I_n - \alpha A)^{-1}
\label{eq:transfer}
\end{equation}
for $n \times n$ real symmetric matrix $A$. Given the eigenvalue decomposition $\cW = U \Lambda U^\T$ of the symmetric matrix $\cW$, Iscen \etal~\cite{IAT+18} observe that $h_\alpha(\cW) = U h_\alpha(\Lambda) U^\T$, so that~(\ref{eq:temporal}) can be written as
\begin{equation}
	\vx = U h_\alpha(\Lambda) U^\T \vy,
\label{eq:spectral}
\end{equation}
which they approximate by keeping only the largest $r$ eigenvalues and the corresponding eigenvectors of $\cW$. This defines a low-rank approximation $h_\alpha(\cW) \approx U_1 h_\alpha(\Lambda_1) U_1^\T$ instead. This method is referred to as \emph{fast spectral ranking} (FSR) in~\cite{IAT+18}. Crucially, $\Lambda$ is a diagonal matrix, hence $h_\alpha$ applies element-wise, as a scalar function $h_\alpha(x) \defn (1 - \alpha) / (1 - \alpha x)$ for $x \in [-1,1]$.

At the expense of off-line computing and storing the $n \times r$ matrix $U_1$ and the $r$ eigenvalues in $\Lambda_1$, filtering is now computed as a sequence of low-rank matrix-vector multiplications, and the query is accelerated by orders of magnitude compared to~\cite{ITA+17}. However, the space overhead is significant. To deal with approximation errors when $r$ is small, a heuristic is introduced that gradually falls back to the similarity in the original space for items that are poorly represented, referred to as FSRw~\cite{IAT+18}.

We call this method \emph{spectral filtering} because $U$ represents the Fourier basis of the graph and the sequence of matrix-vector multiplications from right to left in the right-hand side of~(\ref{eq:spectral}) represents the Fourier transform of $\vy$ by $U^\T$, filtering in the frequency domain by $h_\alpha(\Lambda)$, and finally the inverse Fourier transform to obtain $\vx$ in the time domain by $U$.
\section{Hybrid spectral-temporal filtering}
\label{sec:hybrid}

Temporal filtering~(\ref{eq:temporal}) is performed once for every new query represented by $\vy$, but $\cW$ represents the dataset and is fixed. Could CG be accelerated if we had some very limited additional information on $\cW$?

On the other extreme, spectral filtering~(\ref{eq:spectral}) needs a large number of eigenvectors and eigenvalues of $\cW$ to provide a high quality approximation, but always leaves some error. Could we reduce this space requirement by allocating some additional query time to recover the approximation error?

The answer is positive to both questions and in fact these are the two extreme cases of \emph{hybrid spectral-temporal filtering}, which we formulate next.

\subsection{Derivation}
\label{sec:deriv}

We begin with the eigenvalue decomposition $\cW = U \Lambda U^\T$, which we partition as $\Lambda = \diag(\Lambda_1, \Lambda_2)$ and $U = (U_1 \ U_2)$. Matrices $\Lambda_1$ and $\Lambda_2$ are diagonal $r \times r$ and $(n-r) \times (n-r)$, respectively. Matrices $U_1$ and $U_2$ are $n \times r$ and $n \times (n-r)$, respectively, and have the following orthogonality properties, all due to the orthogonality of $U$ itself:
\begin{equation}
	U_1^\T U_1 = I_r, \quad U_2^\T U_2 = I_{n-r}, \quad
	U_1^\T U_2 = \vO, \quad U_1 U_1^\T + U_2 U_2^\T = I_n.
\label{eq:ortho}
\end{equation}
Then, $\cW$ is decomposed as
\begin{equation}
	\cW = U_1 \Lambda_1 U_1^\T + U_2 \Lambda_2 U_2^\T.
\label{eq:w-decomp}
\end{equation}
Similarly, $h_\alpha(\cW)$ is decomposed as
\begin{align}
	h_\alpha(\cW)
		& = U h_\alpha(\Lambda) U^\T
			\label{eq:h-decomp-1} \\
		& = U_1 h_\alpha(\Lambda_1) U_1^\T + U_2 h_\alpha(\Lambda_2) U_2^\T,
			\label{eq:h-decomp-2}
\end{align}
which is due to the fact that diagonal matrix $h_\alpha(\Lambda)$ is obtained element-wise, hence decomposed as $h_\alpha(\Lambda) = \diag(h_\alpha(\Lambda_1), h_\alpha(\Lambda_2))$. Here the first term is exactly the low-rank approximation that is used by spectral filtering, and the second is the approximation error
\begin{align}
	e_\alpha(\cW)
		& \defn U_2 h_\alpha(\Lambda_2) U_2^\T
			\label{eq:error-1} \\
		& = (1 - \alpha) \left(
				U_2 (I_{n-r} - \alpha \Lambda_2)^{-1} U_2^\T + U_1 U_1^\T - U_1 U_1^\T
			\right)
			\label{eq:error-2} \\
		& = (1 - \alpha) \left(
				\left(U_2 (I_{n-r} - \alpha \Lambda_2) U_2^\T + U_1 U_1^\T \right)^{-1}
				- U_1 U_1^\T
			\right)
			\label{eq:error-3} \\
		& = (1 - \alpha) \left(
				\left( I_n - \alpha U_2 \Lambda_2 U_2^\T \right)^{-1} - U_1 U_1^\T
			\right)
			\label{eq:error-4} \\
		& = h_\alpha(U_2 \Lambda_2 U_2^\T) - (1 - \alpha) U_1 U_1^\T.
			\label{eq:error-5}
\end{align}
We have used the definition~(\ref{eq:transfer}) of $h_\alpha$ in~(\ref{eq:error-2}) and~(\ref{eq:error-5}). Equation~(\ref{eq:error-4}) is due to the orthogonality properties~(\ref{eq:ortho}). Equation~(\ref{eq:error-3}) follows from the fact that for any invertible matrices $A$, $B$ of conformable sizes,
\begin{equation}
	\left( U_1 A U_1 + U_2 B U_2 \right)^{-1} = U_1 A^{-1} U_1 + U_2 B^{-1} U_2,
\label{eq:inversion}
\end{equation}
which can be verified by direct multiplication, and is also due to orthogonality.

Now, combining~(\ref{eq:h-decomp-2}),~(\ref{eq:error-5}) and~(\ref{eq:w-decomp}), we have proved the following.

\begin{theorem}
	Assuming the definition~(\ref{eq:transfer}) of transfer function $h_\alpha$ and the eigenvalue decomposition~(\ref{eq:w-decomp}) of the symmetrically normalized adjacency matrix $\cW$, $h_\alpha(\cW)$ is decomposed as
	\begin{equation}
		h_\alpha(\cW) = U_1 g_\alpha(\Lambda_1) U_1^\T + h_\alpha(\cW - U_1 \Lambda_1 U_1^\T),
	\label{eq:main}
	\end{equation}
	where
	\begin{equation}
		g_\alpha(A)
			\defn h_\alpha(A) - h_\alpha(\vO)
			= (1 - \alpha) \left( (I_n - \alpha A)^{-1} - I_n \right)
	\label{eq:aux}
	\end{equation}
	for $n \times n$ real symmetric matrix $A$. For $x \in [-1,1]$ in particular, $g_\alpha(x) \defn h_\alpha(x) - h_\alpha(0) = (1 - \alpha) \alpha x / (1 - \alpha x)$.
\end{theorem}

Observe that $\Lambda_2, U_2$ do not appear in~(\ref{eq:main}) and indeed it is only the largest $r$ eigenvalues $\Lambda_1$ and corresponding eigenvectors $U_1$ of $\cW$ that we need to compute. The above derivation is generalized from $h_\alpha$ to a much larger class of functions in appendix~\ref{sec:deriv2}.


\subsection{Algorithm}

\emph{Why is decomposition~(\ref{eq:main}) of $h_\alpha(\cW)$ important?} Because given an observation vector $\vy$ at query time, we can express the ranking vector $\vx$ as
\begin{equation}
	\vx = \vx^s + \vx^t,
\label{eq:hybrid}
\end{equation}
where the first, \emph{spectral}, term $\vx^s$ is obtained by spectral filtering
\begin{equation}
	\vx^s = U_1 g_\alpha(\Lambda_1) U_1^\T \vy,
\label{eq:h-spectral}
\end{equation}
as in~\cite{IAT+18}, where $g_\alpha$ applies element-wise, while the second, \emph{temporal}, term $\vx^t$ is obtained by temporal filtering, that is, solving the linear system
\begin{equation}
	\cL_\alpha(\cW - U_1 \Lambda_1 U_1^\T) \vx^t = \vy,
\label{eq:h-temporal}
\end{equation}
which we do by a few iterations of CG as in~\cite{ITA+17}. The latter is possible because $\cL_\alpha(\cW - U_1 \Lambda_1 U_1^\T)$ is still positive-definite, like $\cL_\alpha(\cW)$.
It's also possible without an explicit dense representation of $U_1 \Lambda_1 U_1^\T$ because CG, like all Krylov subspace methods, only needs \emph{black-box} access to the matrix $A$ of the linear system, that is, a mapping $\vz \mapsto A \vz$ for $\vz \in \real^n$. For system~(\ref{eq:h-temporal}) in particular, according to the definition~(\ref{eq:laplacian}) of $\cL_\alpha$, we use the mapping
\begin{equation}
	\vz \mapsto \left(
			\vz - \alpha \left( \cW \vz - U_1 \Lambda_1 U_1^\T \vz \right)
		\right) / (1 - \alpha),
\label{eq:box}
\end{equation}
where product $\cW \vz$ is efficient because $\cW$ is sparse as in~\cite{ITA+17}, while $U_1 \Lambda_1 U_1^\T \vz$ is efficient if computed right-to-left because $U_1$ is an $n \times r$ matrix with $r \ll n$ and $\Lambda_1$ is diagonal as in~\cite{IAT+18}.

\subsection{Discussion}

\emph{What is there to gain from spectral-temporal decomposition~(\ref{eq:hybrid}) of $\vx$?}

First, since the temporal term~(\ref{eq:h-temporal}) can recover the spectral approximation error, the rank $r$ of $U_1$, $\Lambda_1$ in the spectral term~(\ref{eq:h-spectral}) can be chosen as small as we like. In the extreme case $r = 0$, the spectral term vanishes and we recover temporal filtering~\cite{ITA+17}. This allows efficient computation of only a few eigenvectors/values, even with the Lanczos algorithm rather than the approximate method of~\cite{IAT+18}. Most importantly, it reduces significantly the space complexity, at the expense of query time.
Like spectral filtering, it is possible to \emph{sparsify} $U_1$ to compress the dataset embeddings and accelerate the queries online. In fact, we show in section~\ref{sec:exp} that sparsification is much more efficient in our case.

Second, the matrix $\cW - U_1 \Lambda_1 U_1^\T$ is effectively like $\cW$ with the $r$ largest eigenvalues removed. This improves significantly the condition number of matrix $\cL_\alpha(\cW - U_1 \Lambda_1 U_1^\T)$ in the temporal term~(\ref{eq:h-temporal}) compared to $\cL_\alpha(\cW)$ in the linear system~(\ref{eq:temporal}) of temporal filtering~\cite{ITA+17}, on which the convergence rate depends. In the extreme case $r = n$, the temporal term vanishes and we recover spectral filtering~\cite{IAT+18}. In turn, even with small $r$, this reduces significantly the number of iterations needed for a given accuracy, at the expense of computing and storing $U_1$, $\Lambda_1$ off-line as in~\cite{IAT+18}. The improvement is a function of $\alpha$ and the spectrum of $\cW$, and is quantified in section~\ref{sec:analysis}.

In summary, for a given desired accuracy, we can choose the rank $r$ of the spectral term and a corresponding number of iterations of the temporal term, determining a trade-off between the space needed for the eigenvectors (and the off-line cost to obtain them) and the (online) query time. Such choice is not possible with either spectral or temporal filtering alone: at large scale, the former may need too much space and the latter may be too slow.

\section{Analysis}
\label{sec:analysis}

\emph{How ``easier'' for CG is $\cW - U_1 \Lambda_1 U_1^\T$ in~(\ref{eq:h-temporal}) compared to $\cW$ in~(\ref{eq:temporal})?}

In solving a linear system $A \vx = \vb$ where $A$ is an $n \times n$ symmetric positive-definite matrix, CG generates a unique sequence of iterates $\vx_i$, $i=0,1,\dots$, such that the $A$-norm $\norm{\ve_i}_A$ of the error $\ve_i \defn \vx^* - \vx_i$ is minimized over the Krylov subspace $\cK_i \defn \gen{\vb, A \vb, \dots, A^i \vb}$ at each iteration $i$, where $\vx^* \defn A^{-1} \vb$ is the exact solution and the $A$-norm is defined by $\norm{\vx}_A \defn \sqrt{\vx^\T A \vx}$ for $\vx \in \real^n$.

A well-known result on the rate of convergence of CG that assumes minimal knowledge of the eigenvalues of $A$ states that the $A$-norm of the error at iteration $i$, relative to the $A$-norm of the initial error $\ve_0 \defn \vx^*$, is upper-bounded by~\cite{Tref97}
\begin{equation}
	\frac{\norm{\ve_i}_A}{\norm{\ve_0}_A} \le \phi_i(A) \defn 2 \left(
			\frac{\sqrt{\kappa(A)} - 1}{\sqrt{\kappa(A)} + 1}
		\right)^i,
\label{eq:bound}
\end{equation}
where $\kappa(A) \defn \norm{A} \norm{A^{-1}} = \lambda_1(A) / \lambda_n(A)$ is the $2$-norm \emph{condition number} of $A$, and $\lambda_j(A)$ for $j = 1,\dots,n$ are the eigenvalues of $A$ in descending order.

In our case, matrix $\cL_\alpha(\cW)$ of linear system~(\ref{eq:temporal}) has condition number
\begin{equation}
	\kappa(\cL_\alpha(\cW))
		= \frac{1 - \alpha \lambda_n(\cW)}{1 - \alpha \lambda_1(\cW)}
		= \frac{1 - \alpha \lambda_n(\cW)}{1 - \alpha}.
\label{eq:cond}
\end{equation}
The first equality holds because for each eigenvalue $\lambda$ of $\cW$ there is a corresponding eigenvalue $(1 - \alpha \lambda) / (1 - \alpha)$ of $\cL_\alpha(\cW)$, which is a decreasing function. The second holds because $\lambda_1(\cW) = 1$~\cite{Chun97}.

Now, let $\cW_r \defn \cW - U_1 \Lambda_1 U_1^\T$ for $r = 0,1,\dots,n-1$, where $\Lambda_1$, $U_1$ represent the largest $r$ eigenvalues and the corresponding eigenvectors of $\cW$ respectively. Clearly, $\cW_r$ has the same eigenvalues as $\cW$ except for the largest $r$, which are replaced by zero. That is, $\lambda_1(\cW_r) = \lambda_{r+1}(\cW)$ and $\lambda_n(\cW_r) = \lambda_n(\cW)$. The latter is due to the fact that $\lambda_n(\cW) \le -1 / (n-1) \le 0$~\cite{Chun97}, so the new zero eigenvalues do not affect the smallest one. Then,
\begin{equation}
	\kappa(\cL_\alpha(\cW_r))
		= \frac{1 - \alpha \lambda_n(\cW)}{1 - \alpha \lambda_{r+1}(\cW)}
		\le \kappa(\cL_\alpha(\cW)).
\label{eq:cond-r}
\end{equation}
This last expression generalizes~(\ref{eq:cond}). Indeed, $\cW = \cW_0$. Then, our hybrid spectral-temporal filtering involves CG on $\cL_\alpha(\cW_r)$ for $r \ge 0$, compared to the baseline temporal filtering for $r = 0$. The inequality in~(\ref{eq:cond-r}) is due to the fact that $|\lambda_j(\cW)| \le 1$ for $j = 1,\dots,n$~\cite{Chun97}. Removing the largest $r$ eigenvalues of $\cW$ clearly improves (decreases) the condition number of $\cL_\alpha(\cW_r)$ relative to $\cL_\alpha(\cW)$. The improvement is dramatic given that $\alpha$ is close to $1$ in practice. For $\alpha = 0.99$ and $\lambda_{r+1}(\cW) = 0.7$ for instance, $\kappa(\cL_\alpha(\cW_r)) / \kappa(\cL_\alpha(\cW)) = 0.0326$.

\begin{figure}[b!]
\vspace{-6pt}
\begin{tabular}{cc}
\begin{tikzpicture}
\begin{axis}[
	grid=both,
	width=.45\textwidth,
	height=.4\textwidth,
	enlargelimits=false,
	xlabel={order $j$},
	ylabel={eigenvalue $\lambda_j(\cW)$},
]
\addplot[blue] table{figs/rate/eig.txt};
\addplot[red] coordinates {(300,-1) (300,1)};
\end{axis}
\end{tikzpicture}
&
\begin{tikzpicture}
\begin{axis}[
	width=.55\textwidth,
	height=.4\textwidth,
	enlargelimits=false,
	xlabel={rank $r$ (space)},
	ylabel={iteration $i$ (time)},
]
\addplot[
	contour prepared={
		labels over line,
		label distance=70pt,
		contour label style={font=\tiny},
	},
	contour prepared format=matlab,
] table{figs/rate/contour.txt};
\end{axis}
\end{tikzpicture}
\\
(a) & (b)
\end{tabular}
\vspace{-6pt}
\caption{(a) In descending order, eigenvalues of adjacency matrix $\cW$ of Oxford5k dataset of $n = 5,063$ images with global GeM features by ResNet101
and $k = 50$ neighbors per point (see section~\ref{sec:exp}). Eigenvalues on the left of vertical red line at $j=300$ are the largest $300$ ones, candidate for removal. (b) Contour plot of upper bound $\phi_i(\cL_\alpha(\cW_r))$ of CG's relative error as a function of rank $r$ and iteration $i$ for $\alpha = 0.99$, illustrating the space($r$)-time($i$) trade-off for constant relative error.}
\label{fig:analysis}
\end{figure}
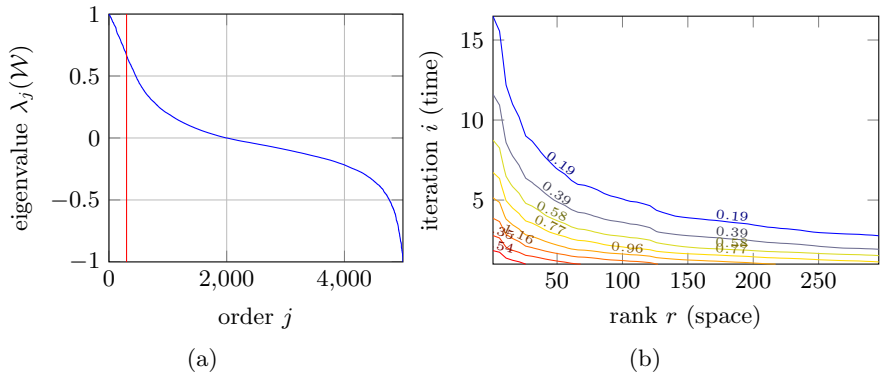

More generally, given the eigenvalues $\lambda_{r+1}(\cW)$ and $\lambda_n(\cW)$, the improvement can be estimated by measuring the upper bound $\phi_i(\cL_\alpha(\cW_r))$ for different $i$ and $r$. A concrete example is shown in Figure~\ref{fig:analysis}, where we measure the eigenvalues of the adjacency matrix $\cW$ of a real dataset, remove the largest $r$ for $0 \le r \le 300$ and plot the upper bound $\phi_i(\cL_\alpha(\cW_r))$ of the relative error as a function of rank $r$ and iteration $i$ given by~(\ref{eq:bound}) and~(\ref{eq:cond-r}). Clearly, as more eigenvalues are removed, less CG iterations are needed to achieve the same relative error; the approximation error represented by the temporal term decreases and at the same time the linear system becomes easier to solve. Of course, iterations become more expensive as $r$ increases; precise timings are given in section~\ref{sec:exp}.

\section{Experiments}
\label{sec:exp}

In this section we evaluate our hybrid method on popular image retrieval benchmarks.
We provide comparisons to baseline methods, analyze the trade-off between runtime complexity, memory footprint and search accuracy, and compare with the state of the art.

\subsection{Experimental setup}
\label{sec:expSetup}

\head{Datasets.}
We use the revisited retrieval benchmark~\cite{RIT+18} of the popular Oxford buildings~\cite{PCISZ07} and Paris~\cite{PCISZ08} datasets, referred to as \roxf and \rpar, respectively.
Unless otherwise specified, we evaluate using the \emph{Medium} setup and always report mean Average Precision (mAP).
Large-scale experiments are conducted on \roxf+\r1m and \rpar+\r1m by adding the new 1M challenging distractor set~\cite{RIT+18}.

\head{Image representation.}
We use GeM descriptors~\cite{RTC18} to represent images.
We extract GeM at 3 different image scales, aggregate the 3 descriptors, and perform whitening, exactly as in~\cite{RTC18}.
Finally, each image is represented by a single vector with $d = 2048$ dimensions, since ResNet-101 architecture is used.

\head{Baseline methods.} We consider the two baseline methods described in Section~\ref{sec:background}, namely temporal and spectral filtering.
\emph{Temporal filtering} corresponds to solving a linear system with CG~\cite{ITA+17} and is evaluated for different numbers of CG iterations. It is used with truncation at large scale to speed up the search~\cite{ITA+17} and is denoted by \emph{Temporal}$\dagger$.
\emph{Spectral filtering} corresponds to FSR and its FSRw variant~\cite{IAT+18}.
Both FSR variants are parametrized by the rank $r$ of the approximation, which is equal to the dimensionality of the spectral embedding.

\head{Implementation details.}
Temporal ranking is performed with the implementation\footnote{\url{https://github.com/ahmetius/diffusion-retrieval/}} provided by Iscen~\etal~\cite{ITA+17}.
The adjacency matrix is constructed by using top $k=50$ reciprocal neighbors.
Pairwise similarity between descriptors $\vv$ and $\vz$ is estimated by $( \vv^{\top}\vz )_+^3$.
Parameter $\alpha$ is set to $0.99$, while the observation vector $\vy$ includes the top $5$  neighbors.
The eigendecomposition is performed on the largest connected component, as in~\cite{IAT+18}.
Its size is 933,412 and 934,809 for \roxf+\r1m and \rpar+\r1m, respectively.
Timings are measured with Matlab implementation on a 4-core Intel Xeon 2.60GHz CPU with 200 GB of RAM.
We only report timings for the diffusion part of the ranking and exclude the initial nearest neighbor search used to construct the observation vector.

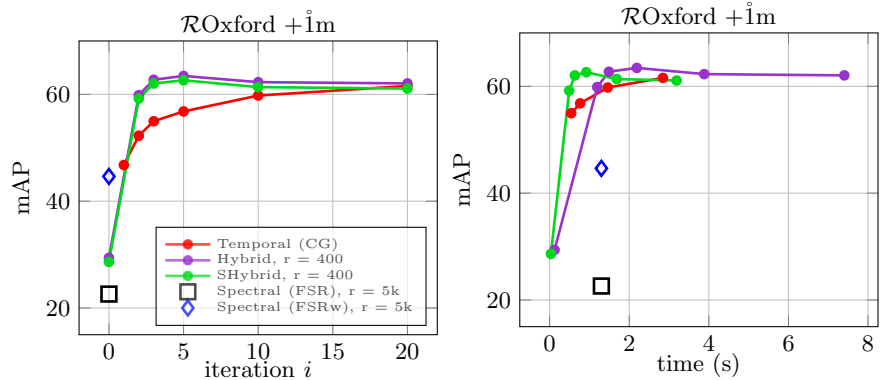
\begin{figure}[t]
\vspace{-5pt}
\definecolor{greeny}{rgb}{0.0,0.85,0.1}
\definecolor{cyany}{rgb}{.6,0.2,0.8}
\begin{center}%
\begin{tabular}{cc}
\begin{tikzpicture}%
\begin{axis}[%
	title = {\roxf+\r1m},
  title style={yshift=-1.7ex},
  xlabel style={yshift=1.4ex},
	width=0.52\linewidth,
	height=0.45\linewidth,
	xlabel={iteration $i$},
	ylabel={mAP},
	grid=both,
	enlarge x limits={0.1},
	xtick distance=5,
  legend style={font=\tiny, fill opacity=0.7, legend pos=south east, legend cell align=left, legend style={row sep=-3pt}},
	xmin = 0, xmax = 20,
	ymin = 15, ymax = 70,
]
	\pgfplotstableread{
		it	r100	r400	rs400 	cg      r100t   r400t  	rs400t 	cgt
		0	0.1498	0.2938 	0.2863	nan		0.0307  0.1248 	0.0319 	nan
		1	nan		nan		nan		0.4677 	nan		nan		nan		0.2279
		2	0.5427	0.5986 	0.5920	0.5226 	0.5894  1.1957 	0.4908 	0.3429
		3	0.5807	0.6274 	0.6204	0.5497  0.8049  1.4895 	0.6336 	0.5417
		5	0.5982	0.6345 	0.6264	0.5681  1.2081  2.1902 	0.9210 	0.7661
		10	0.6059	0.6229 	0.6136	0.5978  2.1555  3.8840 	1.6859 	1.4603
		20	0.619	0.6205 	0.6109	0.6158	4.0706  7.4030 	3.1899 	2.8477
	}{\tabb}
	\addplot[red,    solid, mark=*,  mark size=1.5, line width=1.0]  table[x expr=\thisrow{it},y expr=\thisrow{cg}*100] \tabb; \leg{Temporal (CG)}
	\addplot[cyany,    solid, mark=*,  mark size=1.5, line width=1.0]  table[x expr=\thisrow{it},y expr=\thisrow{r400}*100] \tabb; \leg{Hybrid, r = 400}
	\addplot[greeny, solid, mark=*,  mark options={solid}, mark size=1.5, line width=1.0]  table[x expr=\thisrow{it},y expr=\thisrow{rs400}*100] \tabb; \leg{SHybrid, r = 400}
	\addplot[black,    only marks, mark=square,  mark size=2.8, line width=1.0]  coordinates {(0, 22.61)}; \leg{Spectral (FSR), r = 5k}
	\addplot[blue,    only marks, mark=diamond,  mark size=2.8, line width=1.0]  coordinates {(0, 44.64)}; \leg{Spectral (FSRw), r = 5k}
\end{axis}
\end{tikzpicture}%
&
\begin{tikzpicture}%
\begin{axis}[%
	title = {\roxf+\r1m},
  title style={yshift=-1.7ex},
  xlabel style={yshift=1.4ex},
	width=0.52\linewidth,
	height=0.45\linewidth,
	xlabel={time (s)},
	ylabel={mAP},
	grid=both,
	enlarge x limits={0.1},
  legend style={font=\tiny, fill opacity=0.7, legend pos=south east, legend cell align=left, legend style={row sep=-3pt}},
	xmin = 0, xmax = 7.5,
	ymin = 15, ymax = 70,
]
	\pgfplotstableread{
		it	r100		r400	 rs400 	cg      r100t    r400t  rs400t cgt
		0		0.1498	0.2938 0.2863	nan			0.0307   0.1248 0.0319 nan
		2		0.5427	0.5986 0.5920	nan 		0.5894   1.1957 0.4908 nan
		3		0.5807	0.6274 0.6204	0.5497  0.8049   1.4895 0.6336 0.5417
		5		0.5982	0.6345 0.6264	0.5681  1.2081   2.1902 0.9210 0.7661
		10	0.6059	0.6229 0.6136	0.5978  2.1555   3.8840 1.6859 1.4603
		20	0.619		0.6205 0.6109	0.6158	4.0706   7.4030 3.1899 2.8477
	}{\tabb}
	\addplot[red,    solid, mark=*,  mark size=1.5, line width=1.0]  table[x expr=\thisrow{cgt},y expr=\thisrow{cg}*100] \tabb; 
	\addplot[cyany,    solid, mark=*,  mark size=1.5, line width=1.0]  table[x expr=\thisrow{r400t},y expr=\thisrow{r400}*100] \tabb; 
	\addplot[greeny, solid, mark=*,  mark size=1.5, line width=1.0]  table[x expr=\thisrow{rs400t},y expr=\thisrow{rs400}*100] \tabb; 
	\addplot[black,    only marks, mark=square,  mark size=2.8, line width=1.0]  coordinates {(1.3, 22.61)}; 
	\addplot[blue,    only marks, mark=diamond,  mark size=2.8, line width=1.0]  coordinates {(1.3, 44.64)}; 
\end{axis}
\end{tikzpicture}%
\\[-3pt]
\end{tabular}
\end{center}
\vspace{-10pt}
\caption{mAP \vs CG iteration $i$ and mAP \vs time for temporal, spectral, and hybrid filtering.
Sparsified hybrid is used with sparsity $99$\%.
\label{fig:mainexp}}
\end{figure}

\begin{figure}[t]
\centering
\input{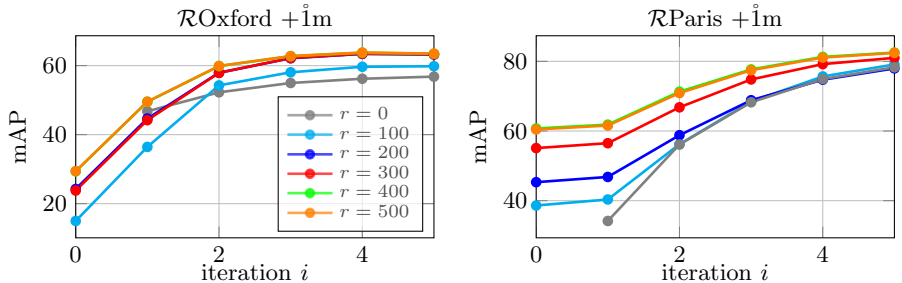}
\small
\begin{tabular}{cc}
{
\begin{tikzpicture}
\begin{axis}[%
	width=0.52\linewidth,
	height=0.35\linewidth,
	xlabel={iteration $i$},
	ylabel={mAP},
	title={\roxf+\r1m},
  title style={yshift=-1.7ex},
  xlabel style={yshift=1.4ex},
	legend pos=south east,
  legend style={cells={anchor=west}, font =\scriptsize, fill opacity=0.7, row sep=-2.5pt},
    xmin = 0, xmax = 5,
    grid=both,
]
	\addplot[gray, solid, mark=*, mark size=1.5, line width=1.0] table[x=it, y expr={100*\thisrow{r0}}] \tradeoffOxft; \leg{$r=0$};
	\addplot[cyan, solid, mark=*, mark size=1.5, line width=1.0] table[x=it, y expr={100*\thisrow{r100}}] \tradeoffOxft;\leg{$r = 100$};
	\addplot[blue, solid, mark=*, mark size=1.5, line width=1.0] table[x=it, y expr={100*\thisrow{r200}}] \tradeoffOxft; \leg{$r = 200$};
	\addplot[red, solid, mark=*, mark size=1.5, line width=1.0] table[x=it, y expr={100*\thisrow{r300}}] \tradeoffOxft; \leg{$r = 300$};
	\addplot[green, solid, mark=*, mark size=1.5, line width=1.0] table[x=it, y expr={100*\thisrow{r400}}] \tradeoffOxft; \leg{$r = 400$};
	\addplot[orange, solid, mark=*, mark size=1.5, line width=1.0] table[x=it, y expr={100*\thisrow{r500}}] \tradeoffOxft; \leg{$r = 500$};
\end{axis}
\end{tikzpicture}
}
&
{
\begin{tikzpicture}
\begin{axis}[%
	width=0.52\linewidth,
	height=0.35\linewidth,
	xlabel={iteration $i$},
	ylabel={mAP},
	title={\rpar+\r1m},
  title style={yshift=-1.7ex},
  xlabel style={yshift=1.4ex},
	legend pos=south east,
    legend style={cells={anchor=west}, font =\scriptsize, fill opacity=0.7, row sep=-2.5pt},
    xmin = 0, xmax = 5,
    grid=both,
]
	\addplot[cyan, solid, mark=*, mark size=1.5, line width=1.0] table[x=it, y expr={100*\thisrow{r100}}] \tradeoffPart;
	\addplot[blue, solid, mark=*, mark size=1.5, line width=1.0] table[x=it, y expr={100*\thisrow{r200}}] \tradeoffPart; 
	\addplot[red, solid, mark=*, mark size=1.5, line width=1.0] table[x=it, y expr={100*\thisrow{r300}}] \tradeoffPart; 
	\addplot[green, solid, mark=*, mark size=1.5, line width=1.0] table[x=it, y expr={100*\thisrow{r400}}] \tradeoffPart; 
	\addplot[orange, solid, mark=*, mark size=1.5, line width=1.0] table[x=it, y expr={100*\thisrow{r500}}] \tradeoffPart; 
	\addplot[gray, solid, mark=*, mark size=1.5, line width=1.0] table[x=it, y expr={100*\thisrow{r0}}] \tradeoffPart; 
\end{axis}
\end{tikzpicture}
}
\end{tabular}
\vspace{-10pt}
 \caption{mAP \vs CG iteration $i$ for different rank $r$ for our hybrid method, where $r=0$ means temporal only.
 \label{fig:tradeoff} }
\vspace{-10pt}
\end{figure}

\subsection{Comparisons}
\head{Comparison with baseline methods.} We first compare performance, query time and required memory for temporal, spectral, and hybrid ranking.
With respect to the memory, all methods store the initial descriptors, \ie one $2048$-dimensional vector per image.
Temporal ranking additionally stores the sparse regularized Laplacian. 
Spectral ranking stores for each vector an additional embedding of dimensionality equal to rank $r$, which is a parameter of the method.
Our hybrid method stores both the Laplacian and the embedding, but with significantly lower rank $r$.

We evaluate on \roxf+\r1m and \rpar+\r1m with global image descriptors, which  is a  challenging large scale problem, \ie large adjacency matrix, where prior methods fail or have serious drawbacks.
Results are shown in Figure~\ref{fig:mainexp}.
Temporal ranking with CG is
reaching saturation near $20$ iterations as in~\cite{ITA+17}.
Spectral ranking is evaluated for a decomposition of rank $r$ whose computation and required memory are reasonable and feasible on a single machine.
Finally, the proposed hybrid method is evaluated for
rank $r=400$, which is a good compromise of speed and memory, as shown below.

Spectral ranking variants (FSR, FSRw) are not performing well despite requiring about $250\%$ additional memory compared to nearest neighbor search in the original space. Compared to hybrid, more than one order of magnitude higher rank $r$ is required for problems of this scale.
Temporal ranking achieves good performance but at much more iterations and higher query times.
Our hybrid solution provides a very reasonable space-time trade-off.

\head{Runtime and memory trade-off.}
We report the trade-off between number of iterations and rank $r$, representing additional memory, in more detail in Figure~\ref{fig:tradeoff}.
It is shown that the number of iterations to achieve the top mAP decreases as the rank increases.
We achieve the optimal trade-off at $r=400$ where we only need 5 or less iterations.
Note that, the rank not only affects the memory and the query time of the spectral part in a linear manner, but the query time of the temporal part too~(\ref{eq:box}).

\begin{figure}
\centering
\input{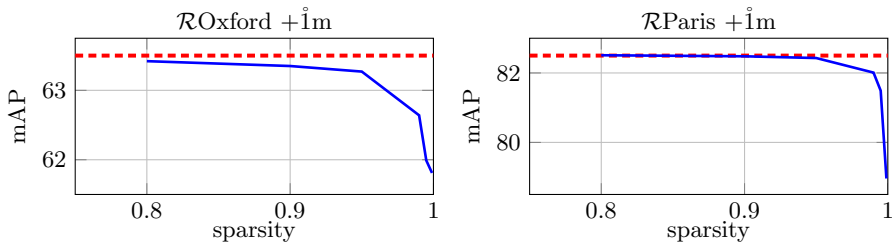}
\small
\begin{tabular}{cc}

\begin{tikzpicture}
\begin{axis}[%
	width=0.52\linewidth,
	height=0.3\linewidth,
	xlabel={sparsity},
	ylabel={mAP},
	title={\roxf+\r1m},
  title style={yshift=-1.7ex},
  xlabel style={yshift=1.4ex},
	legend pos=south west,
    legend style={cells={anchor=west}, font =\scriptsize, fill opacity=0.7, row sep=-2.5pt},
   ymax = 63.75,
   ymin = 61.5,
   xmin = 0.75,
    xmax = 1,
    grid=both,
]
	\addplot[red, densely dashed, line width=1.5] coordinates {(0.75,63.5) (1,63.5)};
	\addplot[blue, solid,  mark size=1.5, line width=1.0] table[x=s, y expr={100*\thisrow{oxf1m}}] \sparse; 
\end{axis}
\end{tikzpicture}
&
\begin{tikzpicture}
\begin{axis}[%
	width=0.52\linewidth,
	height=0.3\linewidth,
	xlabel={sparsity},
	ylabel={mAP},
	title={\rpar+\r1m},
  title style={yshift=-1.7ex},
  xlabel style={yshift=1.4ex},
	legend pos=south west,
    legend style={cells={anchor=west}, font =\scriptsize, fill opacity=0.7, row sep=-2.5pt},
	   ymax = 83,
	   ymin = 78.5,
   	xmin = 0.75,
    xmax = 1,
    grid=both,
   ]
	\addplot[red, densely dashed, line width=1.5] coordinates {(0.75,82.5) (1,82.5)};
	\addplot[blue, solid,  mark size=1.5, line width=1.0] table[x=s, y expr={100*\thisrow{par1m}}] \sparse; 
\end{axis}
\end{tikzpicture}

\end{tabular}
\vspace{-10pt}
 \caption{mAP \vs level of sparsification on our hybrid method for $r = 400$. Dashed horizontal line indicates no sparsification. \label{fig:sparse} }
\end{figure}

\head{Sparsification} of spectral embeddings is exploited in prior work~\cite{IAT+18}.
We sparsify the embeddings of our hybrid method by setting the smallest values of $U_1$ to zero until a desired level of sparsity is reached.
We denote this method by \emph{SHybrid}.
This sparse variant provides memory savings and an additional speedup due to the computations with sparse matrices.
Figure~\ref{fig:sparse} shows that performance loss remains small even for extreme sparsity \eg $99$\%, while the results in Figure~\ref{fig:mainexp} show that it offers a significant speedup in the global descriptor setup.

\begin{table}[t]
\begin{center}
\small
\begin{tabular}{ @{\msp}l@{\msp}@{\msp}c@{\msp}@{\msp}c@{\msp}@{\msp}c@{\msp}@{\msp}c@{\msp}}
      			      &  Temporal~\cite{ITA+17}	    & Temporal$\dagger$~\cite{ITA+17}         & Spectral (FSRw)~\cite{IAT+18}	 & SHybrid        \\ \hline
      mAP		            &  61.6 	            & 59.0                                    & 42.1                     & 62.6               \\
      Time ($s$)	        &  2.8 	            	& 1.0                                    & 1.3                      & 0.9                 \\
      Memory (MB)			&  205 	                & 205                                     & 35,606                      & 264                 \\
\hline
\end{tabular}

\vspace{5pt}
\caption{Performance, memory and query time comparison on \roxf +\r1m with GeM descriptors for temporal (20 iterations), truncated temporal (20 iterations, 75k images in shortlist), spectral ($r=5k$), and hybrid ranking ($r=400$, 5 iterations). Hybrid ranking is sparsified by setting the 99\% smallest values to 0. Reported memory excludes the initial descriptors requiring 8.2 GB. $U_1$ is stored with double precision.
	\label{tab:ptm}}
\end{center}
\end{table}

\begin{figure}[t]
\definecolor{greeny}{rgb}{0.0,0.8,0.2}

\centering%
\input{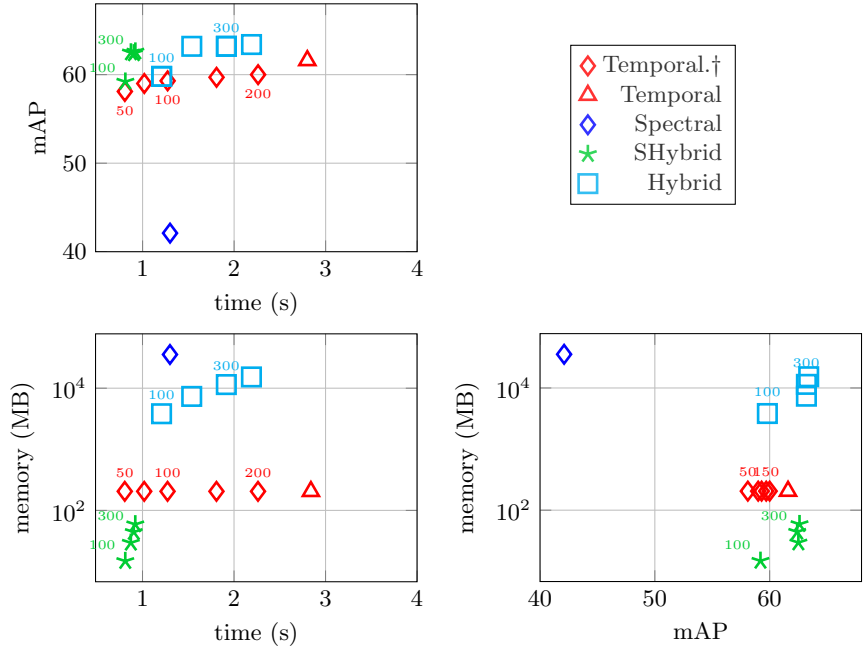}%
\small%
\begin{tabular}{cc}%
%
\multicolumn{2}{l}{
\hspace{7pt}\begin{tikzpicture}%
\begin{axis}[
  width=0.48\linewidth,%
  height=0.4\linewidth,%
  xlabel={time  (s)},%
  ylabel={mAP},%
   legend pos=outer north east,
   legend style={cells={anchor=east}, font =\footnotesize, fill opacity=0.8, row sep=-0.5pt, xshift=14ex, yshift=-4ex},%
   ymax = 68,%
   ymin = 40,%
    xmax = 4.0,%
    grid=both,%
    every node near coord/.append style={font=\tiny},
]%
  \addplot[color=red,    solid, mark=diamond, only marks, mark size=3.5, line width=1.0] table[x=s, y=oxf1m] \truncMemTime; \leg{Temporal.$\dagger$}%
  \addplot[color=red, mark=triangle, only marks, mark size = 3.5, line width = 1] coordinates {(2.8,61.6)};\leg{Temporal}%
  \addplot[color=blue, mark=diamond, only marks, mark size = 3.5, line width = 1] coordinates {(1.3,42.1)};\leg{Spectral}%
  \addplot[color=greeny,    solid, mark=star, only marks, mark size=3.5, line width=1.0] table[x=s, y=map] \shybridoxf; \leg{SHybrid}%
  \addplot[color=cyan,    solid, mark=square, only marks, mark size=3.5, line width=1.0] table[x=s, y=map] \hybridoxf; \leg{Hybrid}%

  \addplot[color=red,    solid, mark=none, only marks, mark size=3.5, line width=1.0, each nth point=2, nodes near coords, nodes near coords style={yshift=-2pt}, nodes near coords align={anchor=north}, point meta=explicit symbolic] table[x=s, y=oxf1m, meta = t] \truncMemTime;
  \addplot[color=greeny,    solid, mark=none, only marks, mark size=3.5, line width=1.0, each nth point=2, nodes near coords, nodes near coords align={anchor=south east}, point meta=explicit symbolic] table[x=s, y=map, meta = r] \shybridoxf;
  \addplot[color=cyan,    solid, mark=square, only marks, mark size=3.5, line width=1.0, each nth point=2,nodes near coords, nodes near coords style={yshift=2pt}, nodes near coords align={anchor=south}, point meta=explicit symbolic] table[x=s, y=map, meta = r] \hybridoxf;
\end{axis}%
\end{tikzpicture}%
}
\\%
\begin{tikzpicture}%
\begin{axis}[
  width=0.48\linewidth,%
  height=0.4\linewidth,%
  xlabel={time (s)},%
  ylabel={memory (MB)},%
  legend pos=north east,%
    legend style={cells={anchor=east}, font =\scriptsize, fill opacity=0.8, row sep=-2.5pt},%
    ymode=log,%
    xmax = 4.0,%
    grid=both,%
    every node near coord/.append style={font=\tiny},
]%
  \addplot[color=red,    solid, mark=diamond, only marks, mark size=3.5, line width=1.0] table[x=s, y=mem] \truncMemTime; 
  \addplot[color=red, mark=triangle, only marks, mark size = 3.5, line width = 1] coordinates {(2.84,205)};
  \addplot[color=blue, mark=diamond, only marks, mark size = 3.5, line width = 1] coordinates {(1.3,35606)};
  \addplot[color=greeny,    solid, mark=star, only marks, mark size=3.5, line width=1.0] table[x=s, y=mem] \shybridoxf; 
  \addplot[color=cyan,    solid, mark=square, only marks, mark size=3.5, line width=1.0] table[x=s, y=mem] \hybridoxf; 

  \addplot[color=red,    solid, mark=none, only marks, mark size=3.5, line width=1.0, each nth point=2,nodes near coords, nodes near coords style={yshift=2pt}, nodes near coords align={anchor=south}, point meta=explicit symbolic] table[x=s, y=mem, meta=t] \truncMemTime; 
  \addplot[color=greeny,    solid, mark=none, only marks, mark size=3.5, line width=1.0, each nth point=2,nodes near coords, nodes near coords style={yshift=1pt}, nodes near coords align={anchor=south east}, point meta=explicit symbolic] table[x=s, y=mem, meta=r] \shybridoxf; 
  \addplot[color=cyan,    solid, mark=none, only marks, mark size=3.5, line width=1.0, each nth point=2,nodes near coords, nodes near coords style={yshift=2pt}, nodes near coords align={anchor=south}, point meta=explicit symbolic] table[x=s, y=mem, meta=r] \hybridoxf; 

\end{axis}%
\end{tikzpicture}%
&%
\raisebox{3pt}{
\begin{tikzpicture}%
\begin{axis}[
  width=0.48\linewidth,%
  height=0.4\linewidth,%
  ylabel={memory (MB)},%
  xlabel={mAP},%
  legend pos=south west,%
  legend style={cells={anchor=east}, font =\scriptsize, fill opacity=0.8, row sep=-2.5pt},%
  xmax = 68, xmin = 40,%
  ymode=log,%
  grid=both,%
  every node near coord/.append style={font=\tiny},
]%
  \addplot[color=red,    solid, mark=diamond, only marks, mark size=3.5, line width=1.0] table[y=mem, x=oxf1m] \truncMemTime; 
  \addplot[color=red, mark=triangle, only marks, mark size = 3.5, line width = 1] coordinates {(61.6,205)};
  \addplot[color=blue, mark=diamond, only marks, mark size = 3.5, line width = 1] coordinates {(42.1,35606)};
  \addplot[color=greeny,    solid, mark=star, only marks, mark size=3.5, line width=1.0] table[y=mem, x=map] \shybridoxf; 
  \addplot[color=cyan,    solid, mark=square, only marks, mark size=3.5, line width=1.0] table[y=mem, x=map] \hybridoxf; 

  \addplot[color=red,    solid, mark=none, only marks, mark size=3.5, line width=1.0, each nth point=3,nodes near coords, nodes near coords style={yshift=2pt}, nodes near coords align={anchor=south}, point meta=explicit symbolic] table[y=mem, x=oxf1m, meta=t] \truncMemTime; 
  \addplot[color=greeny,    solid, mark=none, only marks, mark size=3.5, line width=1.0, each nth point=2,nodes near coords, nodes near coords style={yshift=1pt}, nodes near coords align={anchor=south east}, point meta=explicit symbolic] table[y=mem, x=map, meta=r] \shybridoxf; 
  \addplot[color=cyan,    solid, mark=none, only marks, mark size=3.5, line width=1.0, each nth point=2,nodes near coords, nodes near coords style={yshift=3pt}, nodes near coords align={anchor=south}, point meta=explicit symbolic] table[y=mem, x=map, meta=r] \hybridoxf; 
\end{axis}%
\end{tikzpicture}%
}
\\%
\end{tabular}%
\vspace{-5pt}
\caption{Time (s) - memory (MB) - performance
(mAP) for different methods.
We show mAP \vs time, memory \vs time, and memory \vs mAP on \roxf+\r1m.
Methods in the comparison: temporal for 20 iterations, truncated temporal for 20 iterations and shortlist of size 50k, 75k, 100k, 200k and 300k, spectral (FSRw) with $r=5k$, hybrid with $r \in \{100, 200, 300, 400\}$ and 5 iterations, sparse hybrid with 99\% sparsity, $r \in \{100, 200, 300, 400\}$ and 5 iterations.
Text labels indicate the shortlist size (in thousands) for truncated temporal and rank for hybrid. Observe that the two plots on the left are aligned horizontally with respect to time, while the two at the bottom vertically with respect to memory.
 \label{fig:large_trunc} }
\end{figure}

\head{Performance-memory-speed comparison}
with the baselines is shown in Table~\ref{tab:ptm}.
Our hybrid approach enjoys query times lower than those of temporal with truncation or spectral with FSRw, while at the same time achieves higher performance and requires less memory than the spectral-only approach.

We summarize our achievement in terms of mAP, required memory, and query time in Figure~\ref{fig:large_trunc}.
Temporal ranking achieves high performance at the cost of high query time and its truncated counterpart saves query time but sacrifices performance.
Spectral ranking is not effective at this scale, while our hybrid solution achieves high performance at low query times.

\head{Comparison with the state of the art.}
We present an extensive comparison with existing methods in the literature for global descriptors at small and large scale  (1M distractors).
We choose $r=400$ and $5$ iterations for our hybrid method, $20$ iterations for temporal ranking, $r=2k$ and $r=5k$ for spectral ranking at small and large scale, respectively.
Temporal ranking is also performed with truncation on a shortlist size of $75k$ images  at large scale.
The comparison is presented in Table~\ref{tab:soa}.
Our hybrid approach performs the best or second best right after the temporal one, while providing much smaller query times at a small amount of additional required memory.

\begin{table}[t]
\vspace{10pt}
\begin{center}
\small
\begin{tabular}{ @{\ssp}l@{\ssp}@{\ssp}c@{\ssp}@{\ssp}c@{\msp}@{\msp}c@{\ssp}@{\ssp}c@{\msp}@{\msp}c@{\ssp}@{\ssp}c@{\msp}@{\msp}c@{\ssp}@{\ssp}c@{\ssp}}
      &	\multicolumn{2}{c@{\ssp}}{\roxf}  	& \multicolumn{2}{c@{\ssp}}{\rox+\r1m}	& \multicolumn{2}{c@{\ssp}}{\rpar} &  \multicolumn{2}{c}{\rpar+\r1m} \\
      			                  &  Medium	& Hard	& Medium	& Hard		& Medium		      & Hard	& Medium		      & Hard		\\ \hline
      NN-search		                  &  64.7 	& 38.5      & 45.2 	& 19.9		& 77.2			& 56.3      & 52.3			& 24.7		\\
      $\alpha$-QE~\cite{RIT+18}	      &  67.2 	& 40.8      & 49.0 	& 24.2		& 80.7			& 61.8      & 58.0			& 31.0		\\
      Temporal~\cite{ITA+17}		      &  69.9 	& 40.4      & 61.6 	& 33.2		& 88.9			& 78.5      & 85.0			& 71.6		\\
      Temporal$\dagger$~\cite{ITA+17}	&  - 		& -      	& 59.0 	& 31.4		& -		            & -      	& 79.7			& 65.2		\\
      FSR~\cite{IAT+18}			      &  70.4 	& 42.0      & 22.6 	& 5.8		      & 88.6			& 77.9      & 66.6			& 40.2		\\
      FSRw~\cite{IAT+18}		      &  70.7 	& 42.2      & 42.1 	& 18.8		& 88.7			& 78.0      & 77.4			& 59.7		\\
      SHybrid (ours)			      &  70.5	& 40.3	& 62.6	& 34.4		& 88.7			& 78.1	& 82.0			& 66.8		\\ 
\hline
\end{tabular}

\caption{mAP comparison with existing methods in the literature on small and large scale datasets, using Medium and Hard setup of the revisited benchmark.
	\label{tab:soa}
}
\end{center}
\end{table}

\section{Conclusions} \label{sec:conclusions}

In this work we have tested the two most successful manifold ranking methods of temporal filtering~\cite{ITA+17} and spectral filtering~\cite{IAT+18} on the very challenging new benchmark of Oxford and Paris datasets~\cite{RIT+18}. It is the first time that such methods are evaluated at the scale of one million images. \emph{Spectral filtering}, with both its FSR and FSRw variants, fails at this scale, despite the significant space required for additional vector embeddings. It is possible that a higher rank would work, but it wouldn't be practical in terms of space. In terms of query time, \emph{temporal filtering} is only practical with its truncated variant at this scale. It works pretty well in terms of performance, but the query time is still high.

Our new \emph{hybrid filtering} method allows for the first time to strike a reasonable balance between the two extremes. Without truncation, it outperforms temporal filtering while being significantly faster, and its memory overhead is one order of magnitude less than that of spectral filtering. Unlike spectral filtering, it is possible to extremely sparsify the dataset embeddings with only negligible drop in performance. This, together with its very low rank, makes our hybrid method even faster than spectral, despite being iterative. More importantly, while previous methods were long known in other fields before being applied to image retrieval, to our knowledge our hybrid method is novel and can apply \eg to any field where graph signal processing applies and beyond. Our theoretical analysis shows exactly why our method works and quantifies its space-time-accuracy trade-off using simple ideas from numerical linear algebra.

\footnotesize{
\head{Acknowledgements:} This work was supported by MSMT LL1303 ERC-CZ grant and the OP VVV funded project 
CZ.02.1.01/0.0/0.0/16\_019/0000765 
``Research Center for Informatics''.
}

\appendix
\section{General derivation}
\label{sec:deriv2}

The derivation of our algorithm in section~\ref{sec:deriv} applies only to the particular function (filter) $h_\alpha$~(\ref{eq:transfer}). Here, as in~\cite{IAT+18}, we generalize to a much larger class of functions, that is, any function $h$ that has a series expansion
\begin{equation}
	h(A) = \sum_{i=0}^\infty c_i A^i.
\label{eq:series}
\end{equation}
We begin with the same eigenvalue decomposition~(\ref{eq:w-decomp}) of $\cW$ and, assuming that $h(\cW)$ converges absolutely, its corresponding decomposition
\begin{equation}
	h(\cW) = U_1 h(\Lambda_1) U_1^\T + U_2 h(\Lambda_2) U_2^\T,
\label{eq:2-h-decomp}
\end{equation}
similar to~(\ref{eq:h-decomp-2}), where $U_1$, $U_2$ have the same orthogonality properties~(\ref{eq:ortho}).

Again, the first term is exactly the low-rank approximation that is used by spectral filtering, and the second is the approximation error
\begin{align}
	e_\alpha(\cW)
		& \defn U_2 h(\Lambda_2) U_2^\T
			\label{eq:2-error-1} \\
		& = \sum_{i=0}^\infty c_i U_2 \Lambda_2^i U_2^\T
			\label{eq:2-error-2} \\
		& = \sum_{i=0}^\infty c_i \left( U_2 \Lambda_2 U_2^\T \right)^i - c_0 U_1 U_1^\T
			\label{eq:2-error-3} \\
		& = h(U_2 \Lambda_2 U_2^\T) - h(0) U_1 U_1^\T.
			\label{eq:2-error-4}
\end{align}
Again, we have used the series expansion~(\ref{eq:series}) of $h$ in~(\ref{eq:2-error-2}) and~(\ref{eq:2-error-4}). Now, equation~(\ref{eq:2-error-3}) is due to the fact that
\begin{equation}
	(U_2 \Lambda_2 U_2^\T)^i = U_2 \Lambda_2^i U_2^\T
\label{eq:term-pos}
\end{equation}
for $i \ge 1$, as can be verified by induction, while for $i = 0$,
\begin{equation}
	U_2 \Lambda_2^0 U_2^\T = U_2 U_2^\T = I_n - U_1 U_1^\T = (U_2 \Lambda_2 U_2^\T)^0 - U_1 U_1^\T.
\label{eq:term-zero}
\end{equation}
In both~(\ref{eq:term-pos}) and~(\ref{eq:term-zero}) we have used the orthogonality properties~(\ref{eq:ortho}).

Finally, combining~(\ref{eq:2-h-decomp}),~(\ref{eq:2-error-4}) and~(\ref{eq:w-decomp}), we have proved the following.

\begin{theorem}
	Assuming the series expansion~(\ref{eq:series}) of transfer function $h$ and the eigenvalue decomposition~(\ref{eq:w-decomp}) of the symmetrically normalized adjacency matrix $\cW$, and given that $h(\cW)$ converges absolutely, it is decomposed as
	\begin{equation}
		h(\cW) = U_1 g(\Lambda_1) U_1^\T + h(\cW - U_1 \Lambda_1 U_1^\T),
	\label{eq:2-main}
	\end{equation}
	where
	\begin{equation}
		g(A) \defn h(A) - h(\vO)
	\label{eq:2-aux}
	\end{equation}
	for $n \times n$ real symmetric matrix $A$. For $h = h_\alpha$ and for $x \in [-1,1]$ in particular, $g_\alpha(x) \defn h_\alpha(x) - h_\alpha(0) = (1 - \alpha) \alpha x / (1 - \alpha x)$.
\end{theorem}

This general derivation explains where the general definition of function $g$~(\ref{eq:2-aux}) is coming from in~(\ref{eq:aux}) corresponding to our treatment of $h_\alpha$ in section~\ref{sec:deriv}.

\bibliographystyle{splncs04}
\bibliography{egbib}

\end{document}